\documentstyle[11pt]{article}

\title{On Image Registration and Subpixel Estimation}

\author{Serap A.~Savari \\
Texas A\&M University \\ %Mail Stop 3128 TAMU \\
College Station, TX 77843-3128, USA}
\begin{document}

\maketitle

\begin{abstract}
Image registration is a classical problem in machine vision which
seeks methods to align discrete images of the same scene to subpixel
accuracy in general situations.  As with all estimation problems, the
underlying difficulty is the partial information available about the
ground truth.  We consider a basic and idealized one-dimensional image
registration problem motivated by questions about measurement and about
quantization, and we demonstrate that the extent to which subinterval/subpixel
inferences can be made in this setting depends on a type of complexity
associated with the function of interest, the relationship between the
function and the pixel size, and the number of distinct sampling count
observations available.
\end{abstract}

%{\noindent \footnotesize\textbf{*}Serap A.~Savari, \linkable{savari@tamu.edu} }

%\begin{spacing}{1}   

\section{Introduction}
The correlation of different discrete images of a setting is an extensively
studied problem for fifty years \cite{anuta, tong} with diverse applications
in numerous fields.  The review article Ref.~2 highlights three main objectives
for these investigations.  First, there is an interest in similarity measures
between images.  Second, one may wish to match up a pair of images to achieve
global registration.  Third, there is sometimes an interest in displacement
or translation estimation.  The main focus of Ref.~2 is on applications of the
Fourier-Mellin transform.  However, other techniques like wavelets have
also been considered \cite{wavelets}.

This paper is motivated in part by literature in different areas.  There is
a precise and well-known theorem from digital communications stating that
about $2 \tau W$ sample points are needed to provide a good approximation
to a very long segment of duration $\tau$ of a waveform bandlimited to
frequencies with magnitude at most $W$. \cite{dof}  Therefore, for that
problem the complexity of the function of interest is important in planning
how much information is needed to understand it.  Returning to image
registration, the authors of \cite[p.~4074]{tong} acknowledge that ``different
tasks and scenarios have different requirements.  In addition to the methods,
further improvements should also concentrate on developing advanced 
task-oriented matching frameworks.''  In particular, data collection and
computational considerations are critical in most image registration
applications.  However, there does not appear to be any theory analogous to
Ref.~4 even for the simplest examples of image registration.  The goal of this
paper is to begin to develop such a theory in an idealized but practically
motivated situation.  Like Ref.~4, the focus here will be on a one-dimensional
problem, but the signal will not be bandlimited.  As we will see, there are
constraints on learning from observations even in the absence of computational
considerations and complications like noise.

The need for measurement dates at least as far back as the early Phoenicians
trading around the Mediterranean Sea \cite[p.~1]{dotson}.
Suppose our measuring instrument has graduated marks separated by a standard
length $T$, and we wish to determine the length of a line segment longer than
$T$.  Further suppose that we cannot precisely control where the ends of the
line segment fall relative to the graduated marks of the measuring device;
for example, the images captured from the scanning electron microscope are
subject to drift. \cite{nist}.  Let us also assume that there is no ambiguity
in the binary decision of whether or not a mark meets the line segment.
If we have a single observation that $n$ marks of the device meet the feature,
then we can only infer that the length of the line segment is at least
$(n-1)T$ and less than $(n+1)T$.  However, assuming that repeated
measurements do not alter the length of the line segment or its remaining
in its entirety within the extent of the measuring device, then a subsequent
measurement could have either one more or one less mark meeting the feature.
In this case, the binary images with different numbers of marks could be
aligned up to a partial order, and the uncertainty about the length of the
line segment would decrease from $2T$ to $T$.  While we may be interested
in also measuring objects that are shorter than $T$, we would need a more accurate
instrument to reliably do so.  Furthermore, additional observations could not
decrease this level of uncertainty in the absence of other prior information
about the line segment and/or the measurement process.  However, next suppose
that within the extent of our measuring device we have two stationary line 
segments each of length at least $T$ separated by a distance which is also 
at least $T$ long.  Now a single observation provides not only an uncertainty
of $2T$ for the lengths of each line segment and the space between them but
also a $2T$ uncertainty about the length of a line segment together with the
space separating it from the other line segment and a $2T$ uncertainty about 
the total length from the beginning of one line segment to the end of the other
one.  As we will describe in Section~2, for this example we can capture at
most four different observations of sampling patterns for the line segment
and the space between them which can be aligned to within a partial ordering
of marks.  In this case the four distinct patterns would reveal relationships
about subinterval information up to one of six partitions of a unit cube.
A need for greater accuracy would require different types of prior information.

Quantization has been part of signal processing practice for decades
\cite{pcm, jayant} and is a different motivation to study the results
of ideal sampling at intervals of length $T$ on piecewise constant functions.
Digital images are generally quantized \cite[\S 2.4]{ip} and have a 
quantization error relative to the original analog signal even in the absence
of other noise or distortions.  Because information about the original analog
signal is lost during quantization, it may be more realistic to view the
quantized signal as the ground truth, and that is the approach of this work.
Unlike the interpolation schemes discussed in Ref.~2, we assume that the
pieces of our piecewise constant function each have length at least $T$.
For ease of alignment we also focus here on spatially-limited signals.
Since our primary focus is on the effect of displacement, we will assume
ideal sampling with no noise or other distortions; our goal is to infer as
much as we can about the lengths of each piece of the signal.

The plan for the rest of the paper is as follows.  In Section~2, we introduce
most of our notation and build intuition by studying the simplest examples in
detail.  In Section~3, we descibe the possible observations corresponding to
a general spatially-limited piecewise constant function (with piece lengths
at least $T$ each) and discuss how the inverse problem relates to partitions
of a unit hypercube.  In Section~4, we make some prefatory remarks about 
generalizing to noisy observations and conclude.

\section{Preliminaries}
Assume a spatially-limited piecewise constant function $g(t)$ with 
$m$ contiguous pieces or regions of lengths $R_1, \ R_2 , \ \dots , \ R_m$
with respective values $g_1, \ g_2, \ \dots , g_m$ defined by
\begin{displaymath}
g(t) \; = \; \left\{ \begin{array}{ll}
g_1, & 0 \leq t < R_1 \\
g_i, & \sum_{j=1}^{i-1} R_j  \leq t < \sum_{j=1}^{i} R_j, \; i \geq 2  \\
0, & \mbox{otherwise};
\end{array}
\right.
\end{displaymath}
here $g_1 \neq 0, \; g_m \neq 0,$ and $g_i \neq g_{i+1}, i \in \{1, \ \dots ,
\ m-1\}$.  The sampling interval is known to be of length $T$, and we represent
region length $R_i$ as
\begin{displaymath}
R_i = (n_i-f_i)T,
\end{displaymath}
where for each $i, \; n_i$ is an integer that is at least two and
$0 < f_i \leq 1$; at least one sample is taken from each region.
The exact number $\eta_i$ of samples from region $i$ depends on the placement
of the sampling grid.  Define the offset $\Delta_i \in [0, \ T)$ as the
distance from the left endpoint of region $i$ to the first sampling
point within the region.  We have the following basic result.

\noindent {\bf Proposition 1:} The number $\eta_i$ of samples from region 
$i, \; i \in \{1, \ \dots , \ m\}$, satisfies
\begin{displaymath}
\eta_i \; = \; \left\{ \begin{array}{ll}
n_i, & 0 \leq \Delta_i < (1-f_i )T \\
n_i - 1, & (1-f_i )T \leq \Delta_i < T .
\end{array}
\right.
\end{displaymath}
\noindent {\bf Proof:} If we start counting samples from a sample point at
distance $\Delta_i$ from the left endpoint of region $i$, then the location
of sample $n_i$ occurs at $\Delta_i + (n_i -1) T$
while the length of $R_i$ is $(n_i-f_i)T$ with $0 < f_i \leq 1. \; \Box$

Proposition~1 indicates that if within our set ${\cal S}$ of observations
of the sample counts \\ $( \eta_1 , \eta_2 , \ \dots , \eta_m )$ within the
support of $g(t)$ we have at least one with $\eta_i = n_i$
and one with $\eta_i = n_i - 1$,
then we know the value of $n_i$.  When $R_i$ is an integer multiple of 
$T, \; \eta_i$ can only take on one value and $n_i$ would be unclear without
other prior information.  Proposition~1 also demonstrates that studying
$\eta_i, \; i \in \{1, \ \dots , \ m\}$, in isolation involves a subdivision
of offset values $\Delta_i$ in $[0, \ T)$ into at most two parts with a
boundary depending on $f_i$.  We will see further subdivisions of offset
intervals when we study multiple regions together.

When $m \geq 2$, the set ${\cal S}$ may offer certain types of subinterval
information based on Propostion~1 and the following relationships among
successive offsets.

\noindent {\bf Lemma 2:}  Suppose $m \geq 2$,  The offsets
$\Delta_i, \; i \in \{2, \ \dots , \ m\}$, satisfy
\begin{displaymath}
\Delta_i = (\Delta_{i-1} + f_{i-1} T) \bmod T.
\end{displaymath}
\noindent {\bf Proof:} For convenience, define $R_0 = 0$.  For $i \geq 2$,
the location of the final sample point of region $i-1$ is at
\begin{displaymath}
\sum_{j=0}^{i-2} R_j  + \Delta_{i-1} + (\eta_{i-1} -1) T,
\end{displaymath}
so the location of the first sample point of region $i$ is at
\begin{displaymath}
\sum_{j=0}^{i-2} R_j  + \Delta_{i-1} + \eta_{i-1} T.
\end{displaymath}
Therefore,
\begin{eqnarray*}
\Delta_i & = & \left( \sum_{j=0}^{i-2} R_j  + \Delta_{i-1} + \eta_{i-1} T 
\right) - \sum_{j=0}^{i-1} R_j \\
& = & (\Delta_{i-1} + (\eta_{i-1} - n_{i-1} )T + f_{i-1} T),
\end{eqnarray*}
which implies the result. $\Box$

We next show

\noindent {\bf Proposition 3:} When $m \geq 2$, the pair 
$(\eta_i , \ \eta_{i+1} ), \; i \in \{1, \ \dots , \ m-1\}$, of sample counts 
from regions $i$ and $i+1$ satisfies one of three cases: \\
\noindent Case 1:  If $0 < f_i < 1$ and $f_i + f_{i+1} \leq 1$, then
\begin{displaymath}
(\eta_i , \ \eta_{i+1}) \; = \; \left\{ \begin{array}{ll}
(n_i , \ n_{i+1}), & 0 \leq \Delta_i < (1-f_i -f_{i+1})T \\
(n_i , \ n_{i+1}-1), & (1-f_i -f_{i+1})T \leq \Delta_i < (1-f_i )T \\
(n_i - 1, \ n_{i+1}), & (1-f_i )T \leq \Delta_i < T ;
\end{array}
\right.
\end{displaymath}
\noindent Case 2:  If $0 < f_i < 1$ and $f_i + f_{i+1} \geq 1$, then
\begin{displaymath}
(\eta_i , \ \eta_{i+1}) \; = \; \left\{ \begin{array}{ll}
(n_i , \ n_{i+1}-1), & 0 \leq \Delta_i < (1-f_i )T \\
(n_i - 1 , \ n_{i+1}), & (1-f_i )T \leq \Delta_i < (2-f_i -f_{i+1})T \\
(n_i - 1, \ n_{i+1}-1), & (2-f_i -f_{i+1})T \leq  \Delta_i < T ;
\end{array}
\right.
\end{displaymath}
\noindent Case 3:  If $f_i = 1$, then
\begin{displaymath}
(\eta_i , \ \eta_{i+1}) \; = \; \left\{ \begin{array}{ll}
(n_i - 1 , \ n_{i+1}), & 0 \leq \Delta_i < (1-f_{i+1} )T \\
(n_i - 1, \ n_{i+1}-1), & (1 -f_{i+1})T \leq  \Delta_i < T .
\end{array}
\right.
\end{displaymath}
\noindent {\bf Proof:} For all three cases the values of $\eta_i$ 
immediately follow from Proposition~1,  For Case~1, Lemma~2 implies that when
$(1-f_i -f_{i+1})T \leq \Delta_i < (1-f_i )T$ we have 
$(1-f_{i+1})T \leq \Delta_i < T.$  Similarly,
for Case~2, Lemma~2 implies that when
$(1-f_i )T \leq \Delta_i < (2-f_i -f_{i+1})T$, we have
$0 \leq \Delta_{i+1} < (1-f_{i+1} )T.$  For Case~3, 
$\Delta_{i+1} = \Delta_{i}$.  The values of $\eta_{i+1}$ for all three cases
follow from Proposition~1. $\Box$

Proposition~3 offers a subinterval relationship between region lengths
$R_i$ and $R_{i+1}$ when there are three distinct sample count pairs
$(\eta_i , \ \eta_{i+1} )$.  If this is not available, then there will be
ambiguity about $n_i, \ n_{i+1}$ and/or $f_i + f_{i+1}$ that can only be
resolved with other prior information.

The next result is key to understanding the possible sample counts
for a set of $K$ contiguous regions.  Let $\lceil x \rceil$ denote the
least integer greater than or equal to $x$.  We have 

\noindent {\bf Proposition 4:} The projection
$(\eta_i , \ \eta_{i+1} , \ \dots , \ \eta_{i+k-1} ), \; i \geq 1, \
i+k-1 \leq m$, of $(\eta_1 , \ \dots , \ \eta_m )$
will be defined by at most $k+1$ $\Delta_i$ 
offset intervals $[0, \ \Delta_i^{(1)}), \; 
[\Delta_i^{(1)} , \ \Delta_i^{(2)}), \
\dots ,  [\Delta_i^{(k)} , \ T),$ where the values of $\Delta_i^{(1)} , \ 
\Delta_i^{(2)}, \ \dots , \\  \Delta_i^{(k)}$ are 
\begin{displaymath}
(1-f_i ) T, \;
(\lceil f_i + f_{i+1} \rceil - (f_i + f_{i+1})T, \ \dots , \
\left( \lceil \sum_{j=0}^k  f_{i+j} \rceil - \sum_{j=0}^k  f_{i+j} \right) T
\end{displaymath}
sorted in ascending order. 

\noindent {\bf Proof (by induction):} Propositions 1 and 3 establish the
result for $k=1$ and  $k=2$.  Suppose it is true at $k=K$ and we wish to show
that it continues to hold at $k=K+1.$  By Lemma~2, the offset
$\tilde{\Delta}_{i+K} = (1-f_K ) T$
for region $i+K$ corresponds to the offset
$\tilde{\Delta}_i = 
\left( \lceil \sum_{j=0}^K  f_{i+j} \rceil - \sum_{j=0}^K  f_{i+j} \right) T$
for region $i$.  For $k=K$, there is one offset interval, say
$[\Delta_i^{(h)} , \ \Delta_i^{(h+1)})$ which contains $\tilde{\Delta}_i$.
For $k=K+1$ we will divide $[\Delta_i^{(h)} , \ \Delta_i^{(h+1))}$ into
$[\Delta_i^{(h)} , \ \tilde{\Delta}_i )$ and 
$[\tilde{\Delta}_i , \ \Delta_i^{(h+1)})$ 
and leave the other intervals as they are.  This collection of intervals
meets the requirements on endpoints and cardinality.  If for $k=K$ the 
interval $[\Delta_i^{(h)} , \ \Delta_i^{(h+1)})$ corresponds to the
sample counts $(\eta_i^{(h)} , \ \eta_{i+1}^{(h)} , \ \dots , \ 
\eta_{i+K-1}^{(h)} ),$
then for $k=K+1$ the intervals for $[\Delta_i^{(h)} , \ \tilde{\Delta}_i )$ and 
$[\tilde{\Delta}_i , \ \Delta_i^{(h+1)})$ correspond to the sample counts
$(\eta_i^{(h)} , \ \eta_{i+1}^{(h)} , \ \dots , \ 
\eta_{i+K-1}^{(h)} , \ n_{i+K} ),$ and \\
$(\eta_i^{(h)} , \ \eta_{i+1}^{(h)} , \ \dots , \ 
\eta_{i+K-1}^{(h)} , \ n_{i+K}-1)$.  To show that the remaining offset
intervals for $k=K$ correspond to projections with a unique extension
at $k=K+1$, it suffices to show that the offset $\tilde{\Delta}_{i+K} = 0$
for region $i+K$ corresponds to one of the region $i$ offset endpoints
$\Delta_i^{(1)} , \  \Delta_i^{(2)}, \ \dots ,  \Delta_i^{(K)}$.  By Lemma~2,
\begin{displaymath}
\Delta_{i+K} = (\Delta_{i} + \left( \sum_{j=0}^{K-1} f_{i+j} \right) T) \bmod T.
\end{displaymath}
By the inductive hypothesis, one of $\{ \Delta_i^{(1)} , \  \Delta_i^{(2)}, \ 
\dots ,  \Delta_i^{(K)} \}$ is 
\begin{displaymath}
\left( \lceil \sum_{j=0}^{K-1} f_{i+j} \rceil - \sum_{j=0}^{K-1} f_{i+j} 
\right) T,
\end{displaymath}
which corresponds to ${\Delta}_{i+K} = 0 \bmod T. \; \Box$

If we omit the special cases of regions or clusters of contiguous regions with
length that is an integer multiple of $T$, then for a spatially-limited,
piecewise constant function with $m$ regions there will be $m+1$ possible
distinct observation vectors of sample counts of the support of the function.
If all of them are available, then the subinterval relationships can be
determined up to one of the $m!$ partitions of a unit hypercube.  When
$m=3$, the six sets of four observation vectors and the corresponding
portion of the unit cube are as follows:

\begin{itemize}
\item $\{(n_1, \ n_2 , \ n_3), \ (n_1, \ n_2 , \ n_3 -1), \
(n_1, \ n_2-1 , \ n_3), \ (n_1 -1, \ n_2 , \ n_3) \}$ \\
$f_1 + f_2 + f_3 < 1.$
\item $\{ (n_1, \ n_2 , \ n_3 -1), \ (n_1, \ n_2-1 , \ n_3), \ 
(n_1 -1, \ n_2 , \ n_3) , (n_1 -1, \ n_2 , \ n_3 -1) \}$ \\
$f_1 + f_2 < 1, \; f_2 + f_3 < 1, \; f_1 + f_2 + f_3 > 1.$
\item $\{ (n_1, \ n_2 , \ n_3 -1), \ (n_1, \ n_2-1 , \ n_3), \
(n_1 , \ n_2 -1 , \ n_3 -1) , \\ (n_1 -1, \ n_2 , \ n_3 -1) \}$ \\
$f_1 + f_2 < 1, \; f_2 + f_3 > 1.$
\item $\{ (n_1, \ n_2-1 , \ n_3), \ (n_1 -1, \ n_2 , \ n_3) , \
(n_1 -1, \ n_2 , \ n_3 -1) , \\ (n_1 -1, \ n_2-1 , \ n_3 ) \}$ \\
$f_1 + f_2 > 1, \; f_2 + f_3 < 1.$
\item $\{ (n_1, \ n_2-1 , \ n_3), \ (n_1 , \ n_2 -1 , \ n_3 -1) , \
(n_1 -1, \ n_2 , \ n_3 -1) , \\ (n_1 -1, \ n_2-1 , \ n_3 ) \}$ \\
$f_1 + f_2 > 1, \; f_2 + f_3 > 1, \; f_1 + f_2 + f_3 < 2.$
\item $\{ (n_1 , \ n_2 -1 , \ n_3 -1) , \ (n_1 -1, \ n_2 , \ n_3 -1) , \
 (n_1 -1, \ n_2-1 , \ n_3 ) , \\  (n_1 -1, \ n_2-1 , \ n_3 -1)  \}$ \\
$f_1 + f_2 + f_3 > 2.$
\end{itemize}

In the next section we will apply and extend the preceding results to generate
the collection of observable vectors of sample counts corresponding to
$R_1, \ \dots , \ R_m$.

\section{Main Results}
To simplify the discussion we will henceforth assume that there are no
integers $i$ and $h$ for which $\sum_{j=0}^h f_{i+j}$ is an integer, but 
the set of observable vectors of sample counts in such cases is an
intersection of sets of observable vectors in the cases we consider.
The main idea in the construction of observable vectors of sample counts
is to use Propositions~1 and 3 as base cases and to progressively extend
the $K+1$ projections of $(\eta_1 , \ \dots , \ \eta_m )$ to
$(\eta_i , \ \eta_{i+1} , \ \dots , \ \eta_{i+K-1} ), \; i \geq 1, \
i+K \leq m$ to the set of $K+2$ projections
$(\eta_i , \ \eta_{i+1} , \ \dots , \ \eta_{i+K} ).$  We introduce the
following notation.  Let $\lfloor x \rfloor , \; x \geq 0,$ denote the
integer part of $x$.  For non-negative integers $K$, define
\begin{displaymath}
\kappa (i, \ K) \; = \; \lfloor \sum_{j=0}^{K} f_{i+j}  \rfloor .
\end{displaymath}
If $\kappa (i, \ K) \geq 1,$ then for $h \in \{1, \ \dots , \kappa (i, \ K)\}$
let $\beta_h (i, \ K)$ be the non-negative integer for which
\begin{displaymath}
\sum_{j=\beta_h (i, \ K)+1}^{K} f_{i+j} < h \; 
\mbox{and} \; \sum_{j=\beta_h (i, \ K)}^{K} f_{i+j} > h .
\end{displaymath}
To simplify notation, we will assume in the following that $i$ is fixed
and write $\kappa (K) = \kappa (i, \ K)$ and
$\beta_h (K) = \beta_h (i, \ K)$.
We first identify vectors of sample counts which have two extensions.

\noindent {\bf Theorem 5:} Suppose we have all $K+1$ vectors of sample counts \\
$(\eta_i , \ \eta_{i+1} , \ \dots , \ \eta_{i+K-1} )$ and $i+K \leq m$.
If $\kappa (K) = 0$, then the vector with both extensions
$\eta_{i+K} = n_{i+K} $ and $\eta_{i+K} = n_{i+K} -1$ satisfies
$\eta_{i+j} = n_{i+j}, \; j \in \{0, \ \dots , \ K-1\} $ .
If $\kappa (K) \geq 1$, then the vector with both extensions
$\eta_{i+K} = n_{i+K} $ and $\eta_{i+K} = n_{i+K} -1$ satisfies
\begin{eqnarray*}
\eta_{i+j} = n_{i+j} -1, & & j \in \{ \beta_{\kappa (K)} (K), \ \dots ,
\beta_1 (K) \} \\
\eta_{i+j} = n_{i+j} , & & j \not\in \{ \beta_{\kappa (K)} (K), \ \dots ,
\beta_1 (K) \} 
\end{eqnarray*}
\noindent {\bf Proof:} We begin by showing
\begin{displaymath}
\sum_{j=0}^{K-1} \eta_{i+j} = \left( \sum_{j=0}^{K-1} n_{i+j} \right) 
- {\kappa (K)} .
\end{displaymath}
Since $\sum_{j=0}^{K} R_{i+j} = \left(  \sum_{j=0}^{K} n_{i+j} \right) -
\left( \sum_{j=0}^{K} f_{i+j} \right) $, 
by an extension of Proposition~1, every observation
$(\eta_i^{'} , \ \eta_{i+1}^{'} , \ \dots , \ \eta_{i+K}^{'} )$ 
of sample counts from regions $i$ to $i+K$ of the support of $g(t)$
satisfies
\begin{equation}
\left(  \sum_{j=0}^{K} n_{i+j} \right) - ({\kappa (K)} +1) \leq
\sum_{j=0}^{K} \eta_{i+j}^{'} \leq
\left(  \sum_{j=0}^{K} n_{i+j} \right) - {\kappa (K)} .
\end{equation}
Suppose we consider an observation with
\begin{displaymath}
\sum_{j=0}^{K-1} \eta_{i+j}^{'} \leq
\left(  \sum_{j=0}^{K-1} n_{i+j} \right) - ({\kappa (K)} +1).
\end{displaymath}
Then the extension 
$(\eta_i^{'} , \ \eta_{i+1}^{'} , \ \dots , \ \eta_{i+K-1}^{'} , \ n_{i+K}-1)$ 
satisfies
\begin{displaymath}
\sum_{j=0}^{K} \eta_{i+j}^{'} \leq
\left(  \sum_{j=0}^{K} n_{i+j} \right) - ({\kappa (K)} +2),
\end{displaymath}
which violates (1).  Next consider an observation with
\begin{displaymath}
\sum_{j=0}^{K-1} \eta_{i+j}^{'} \geq
\left(  \sum_{j=0}^{K-1} n_{i+j} \right) - ({\kappa (K)} -1).
\end{displaymath}
Then the extension 
$(\eta_i^{'} , \ \eta_{i+1}^{'} , \ \dots , \ \eta_{i+K-1}^{'} , \ n_{i+K})$ 
satisfies
\begin{displaymath}
\sum_{j=0}^{K} \eta_{i+j}^{'} \geq
\left(  \sum_{j=0}^{K} n_{i+j} \right) - ({\kappa (K)} -1),
\end{displaymath}
which also violates (1).  Hence, for the observation of interest,
\begin{displaymath}
\sum_{j=0}^{K-1} \eta_{i+j} = \left( \sum_{j=0}^{K-1} n_{i+j} \right) 
- {\kappa (K)} .
\end{displaymath}
If ${\kappa (K)} =0$, then by Proposition~1,
 $\eta_{i+j} = n_{i+j}, \; j \in \{0, \ \dots , \ K-1\} $.
To complete the proof, assume ${\kappa (K)}  \geq 1$.
It remains to show that
\begin{displaymath}
\eta_{i+j} = n_{i+j} -1,  j \in \{ \beta_{\kappa (K)} (K), \ \dots ,
\beta_1 (K) \}.
\end{displaymath}
We proceed with an argument by contradiction.  Suppose instead that
there are $b_{\kappa (K)} , \ \dots , b_1  $ 
with $b_{\kappa (K)}  < b_{\kappa (K)-1} < \dots < b_1 $ for which
\begin{displaymath}
\eta_{i+j} = n_{i+j} -1,  j \in \{ b_{\kappa (K)} , \ \dots , b_1  \},
\end{displaymath}
and $(b_{\kappa (K)} , \ \dots , b_1 ) \neq
(\beta_{\kappa (K)} (K), \ \dots , \beta_1 (K))$.
Then there is a smallest index $k \in \{1, \ \dots , \ {\kappa (K)} \}$
for which $b_k \neq \beta_k (K)$.  If $b_k > \beta_k (K)$, then
\begin{displaymath}
k-1 < \sum_{j=b_k}^K f_{i+j} < k.
\end{displaymath}
By an extension of Proposition 1,
\begin{equation}
\left(  \sum_{j=b_k}^{K} n_{i+j} \right) - k \leq
\sum_{j=b_k}^{K} \eta_{i+j} \leq
\left(  \sum_{j=b_k}^{K} n_{i+j} \right) - (k-1).
\end{equation}
If we choose the extension with $\eta_{i+K} = n_{i+K}-1$, then
\begin{displaymath}
\sum_{j=b_k}^{K} \eta_{i+j} =
\left(  \sum_{j=b_k}^{K} n_{i+j} \right) - (k+1),
\end{displaymath}
contradicting (2).  Therefore, $b_k < \beta_k (K)$.
Hence, by an extension of Proposition 1,
\begin{equation}
\left(  \sum_{j=\beta_k (K)}^{K} n_{i+j} \right) - (k+1) \leq
\sum_{j=\beta_k (K)}^{K} \eta_{i+j} \leq
\left(  \sum_{j=\beta_k (K)}^{K} n_{i+j} \right) - k.
\end{equation}
If we choose the extension with $\eta_{i+K} = n_{i+K}$, then
\begin{displaymath}
\sum_{j=\beta_k (K)}^{K} \eta_{i+j} =
\left(  \sum_{j=\beta_k (K)}^{K} n_{i+j} \right) - (k-1),
\end{displaymath}
contradicting (3).  $\Box$

The final step in generating all possible sample counts given region lengths
$R_i, \ \dots , \ R_{i+K}$ is determining how to uniquely extend an
observation $(\eta_i , \ \eta_{i+1} , \ \dots , \ \eta_{i+K-1} )$ of sample
counts which is not covered by Theorem~5.  We have the following result.

\noindent {\bf Theorem 6:} Suppose we have all $K+1$ vectors of sample counts \\
$(\eta_i , \ \eta_{i+1} , \ \dots , \ \eta_{i+K-1} )$. 
\begin{itemize}
\item If $\sum_{j=0}^{K-1} \eta_{i+j} = (\sum_{j=0}^{K-1}  n_{i+j}) -
(\kappa (K) +1)$, then $\eta_{i+K} = n_{i+K} $.
\item If $\sum_{j=0}^{K-1} \eta_{i+j} = (\sum_{j=0}^{K-1}  n_{i+j}) -
(\kappa (K) -1)$, then $\eta_{i+K} = n_{i+K}-1 $.
\item If there are $b_{\kappa (K)} , \ \dots , b_1 $ 
with $b_{\kappa (K)} < b_{\kappa (K)-1} < \dots < b_1 $ for which
\begin{displaymath}
\eta_{i+j} = n_{i+j} -1,  \; j \in \{ b_{\kappa (K)} , \ \dots , b_1  \},
\end{displaymath}
and $(b_{\kappa (K)} , \ \dots , b_1 ) \neq
(\beta_{\kappa (K)} (K), \ \dots , \beta_1 (K))$,
then let \\ $k \in \{1, \ \dots , \ {\kappa (K)} \}$ be the smallest index
for which $b_k \neq \beta_k (K)$.  If $b_k > \beta_k (K)$, then
$\eta_{i+K} = n_{i+K} $.  If $b_k < \beta_k (K)$, then
$\eta_{i+K} = n_{i+K} -1$.
\end{itemize}
\noindent {\bf Proof:} Since 
$\kappa (K) < \sum_{j=0}^K f_{i+j} < \kappa (K) +1,$
it follows from an extension of Proposition~1 that
\begin{displaymath}
\left(  \sum_{j=0}^{K} n_{i+j} \right) - ({\kappa (K)}  +1) \leq
\sum_{j=0}^{K} \eta_{i+j} \leq
\left(  \sum_{j=0}^{K} n_{i+j} \right) - {\kappa (K)} .
\end{displaymath}
Furthermore, either $\kappa (K-1) = \kappa (K)$ or
$\kappa (K-1) = \kappa (K)-1.$  These facts imply the first two cases.
For the final case, if $b_k > \beta_k (K)$, then
$k-1 < \sum_{j=b_k}^K f_{i+j} < k.$  By (2), $\eta_{i+K} = n_{i+K} $.  
Finally, if $b_k < \beta_k (K)$, then (3) implies that
$\eta_{i+K} = n_{i+K} -1. \; \Box$

Observe that if we have the full set of $m+1$ possible observation vectors
of sample counts of the regions in the support of $g(t)$, then Propositions~1
and 3 and Theorems~5 and 6 define $n_i, \ i \in \{1, \ \dots , \ m\},$ and in
which of the $m!$ partitions of the unit hypercube $(f_1, \ \dots , \ f_m)$
resides.  However, we may not have the full set, and then the best we can
do is to learn from projections corresponding to single regions or to 
contiguous sets of regions.  Continuing the discussion following Proposition~3,
if our projections for regions $1$ and $2$ only yield $(n_1, n_{2}-1)$
and $(n_1 -1,\  n_{2})$, then we learn the values of $n_1$ and $n_{2}$,
but we do not obtain the relationship between $f_1$ and $f_{2}$.
If instead we only see $(n_1, n_{2})$ and $(n_1, n_{2}-1)$, then we
learn the value of $n_{2}$, but if there is a third vector to complete
the set, then it could be either $(n_1 -1,\  n_{2})$ or
$(n_{1}+1, n_{2}-1)$, so we cannot infer $n_1$ or
the relationship between $f_1$ and $f_{2}$.  However, sometimes two
observations can provide some subinterval information.  For example,
if $m=3$ and we observe the pair of sample counts $(n_1, n_2-1, n_3)$
and $(n_1-1, n_2, n_3-1)$, then we know $n_1, \ n_2$ and $n_3$; furthermore,
since in one case $\eta_1 + \eta_2 + \eta_3 = n_1+n_2+n_3 -1$ and in
the other case $\eta_1 + \eta_2 + \eta_3 = n_1+n_2+n_3 -2$, then we also
know that $1<f_1+f_2+f_3<2$.  To look at a simlar example for $m=5$,
if we receive the pair of sample counts $(n_1, n_2-1, n_3, n_4-1, n_5)$ and
$(n_1-1, n_2, n_3-1, n_4, n_5-1)$, then we can infer $n_1, \ n_2, \ n_3, \ n_4,
\ n_5, 1 < f_1+f_2+f_3 < 2, \ 1 < f_2+f_3+f_4 < 2, \\ 1 < f_3+f_4+f_5 < 2,$
and $2 < f_1 + f_2 + f_3+f_4+f_5 < 3$.

\section{Conclusions and a Few Remarks on Noise}

We have initiated a new theoretical approach to the study of discrete image
registration.  For the basic and idealized problem that we considered,
we established limits on subinterval/subpixel estimation in the absence of
prior information about the function of interest or the displacement process
as well as the absence of distortions or considerations about the procurement
of sample count data.  However, a brief commentary on noise may be helpful.
First consider the simplest case where all but one image has no noise and we
are given a single image with noise.  Then we know the number of regions of
the support and the value of the function in each region.  Therefore, the
goal is is determine the sample counts for the regions in the support of the
noisy image;  there are finitely many possibilities with constraints on
sample counts, $\sum_{j=0}^K \eta_{i+j}, i \in \{1, \ \dots , \ m\}, \;
K \in \{0, \ \dots , \ m-i\}$ that are consistent with the other data.
Part of the solution involves the segmentation problem of establishing
where the support begins and ends.  The overall problem is a 
detection/hypothesis testing/classification problem, and there is a
large literature on the detection of signals in noise \cite{detection}.
The more realistic problem is global image registration when all of the
images are noisy\cite{nist}.  For example, in semiconductor metrology the
critical dimension scanning electron microscope can interact with 
and alter the measurement sample \cite[p.~11]{ssv2015}, and so there is
a desire to minimize these effects; this may compromise image fidelity,
which is also a source of concern \cite[p.~120]{ssv2015}.  Returning to
our one-dimensional problem where the support of $g(t)$ does not change
over multiple observations apart from translation, it would be ideal to
jointly process the images to obtain a consistent composite belief about
the number of regions $m$ in the support of $g(t)$, the value of the
function in each region, and the bounds on $\sum_{j=0}^K \eta_{i+j}, 
i \in \{1, \ \dots , \ m\}, \; K \in \{0, \ \dots , \ m-i\}$.
Deep learning \cite{nature} approaches may be natural candidates for this
objective.  Finally, we end this paper on alignment by mentioning that
the term ``noise'' has also been applied to problems involving simpler
human judgment\cite{truth}; there are questions in Ref.~13 about the
practice of working with a single ground truth.

\end{document}